\documentclass[conference]{IEEEtran}
\IEEEoverridecommandlockouts
\usepackage{cite}
\usepackage{amsmath,amssymb,amsfonts}
\usepackage{algorithm}
\usepackage{algorithmic}
\usepackage{graphicx}
\usepackage{textcomp}
\usepackage{xcolor}
\usepackage{booktabs}
\usepackage{subcaption}
\def\BibTeX{{\rm B\kern-.05em{\sc i\kern-.025em b}\kern-.08em
    T\kern-.1667em\lower.7ex\hbox{E}\kern-.125emX}}
\begin{document}

\title{GEAR: From Dynamic Encoding to Dynamic Activation in Social Trajectory Prediction\\

}



\author{

\begin{tabular}{ccc}
\begin{tabular}{c}
Jiaheng Chen \\
\textit{Software College} \\
\textit{Northeastern University} \\
Shenyang, China \\
20236778@stu.neu.edu.cn
\end{tabular}
&
\begin{tabular}{c}
Jiaxing Li \\
\textit{Software College} \\
\textit{Northeastern University} \\
Shenyang, China \\
20237096@stu.neu.edu.cn
\end{tabular}
&
\begin{tabular}{c}
Leixia Wang$^*$%
\thanks{$^*$Corresponding Author: Leixia Wang.} \\
\textit{Software College} \\
\textit{Northeastern University} \\
Shenyang, China \\
wangleixia@neu.edu.cn
\end{tabular}
\end{tabular}

\\[1em]

\begin{tabular}{cc}
\begin{tabular}{c}
Jianan Ju \\
\textit{School of Computer Science and Engineering} \\
\textit{Northeastern University} \\
Shenyang, China \\
2290173@stu.neu.edu.cn
\end{tabular}
&
\begin{tabular}{c}
Tinghe Zhang \\
\textit{Software College} \\
\textit{Northeastern University} \\
Shenyang, China \\
zhangtinghe5@gmail.com
\end{tabular}
\end{tabular}
}

\maketitle

\begin{abstract}
Human trajectory prediction requires modeling both individual motion patterns and social interactions among agents. Existing methods have made substantial progress by using attention mechanisms, graph structures, and temporal encoders to capture dynamic social context. However, most of them primarily focus on how social information is encoded, while paying less explicit attention to how the encoded social context should take effect during future trajectory generation. In this paper, we argue that dynamic social encoding does not necessarily imply dynamic social activation. The same interaction context may require different activation strengths across future horizons and scene densities: social cues should be strengthened when interaction evidence is strong, but suppressed when they are weak or noisy. To address this issue, we propose GEAR, a generation-aware bias activation model for human trajectory prediction. Built upon a bias-decomposed trajectory generation formulation, GEAR dynamically activates the individual-motion and social-resonance bias terms at each future step before final trajectory composition. This allows the model to explicitly control when and how strongly individual and social bias components participate in generation. Experiments on ETH-UCY, SDD, and NBA show that GEAR consistently improves the resonance-based baseline and achieves competitive state-of-the-art performance. Further analyses of activation patterns and density-grouped errors validate the importance of calibrating encoded social context during trajectory generation. Our code is available at https://github.com/11isnotavailable/GEAR.git.

\end{abstract}

\begin{IEEEkeywords}
Human trajectory prediction, social interaction modeling, trajectory forecasting, bias composition, generation-aware activation.
\end{IEEEkeywords}

\section{Introduction}

Human trajectory prediction aims to forecast the future positions of agents given their observed motion histories and environmental context, serving as a fundamental task in autonomous driving, robot navigation, surveillance, crowd analysis, and sports analytics.\cite{rudenko_2020_human} The inherent challenge of this task lies in the fact that human motion is not only driven by individual intention and motion continuity, but also shaped by interactions with nearby agents. \cite{alahi_2016_social}\cite{helbing_1995_social} For example, a person might follow a straight trajectory in a sparse environment, yet instinctively decelerate or yield when navigating around peers. Consequently, an effective prediction model should jointly account for individual motion patterns and dynamic social constraints.\cite{mangalam_2020_it}\cite{salzmann_2020_trajectron}

A large body of work has been devoted to modeling social interactions in trajectory prediction. Early methods use recurrent networks and social pooling \cite{alahi_2016_social}\cite{gupta_2018_social} to aggregate neighboring agents, while later studies introduce attention mechanisms \cite{sadeghian_2019_sophie}, graph neural networks \cite{mohamed_2020_socialstgcnn}, Transformers \cite{yuan_2021_agentformer}, and structured interaction encoders to capture richer social context \cite{wang_2025_siat}. More recently, decomposition-based methods attempt to separate different motion factors, such as individual intention, social behavior, and trajectory randomness \cite{chen_2025_socialmoif}\cite{mangalam_2020_it}\cite{zhang_2024_demo}, leading to more interpretable prediction frameworks. In particular, resonance-based trajectory prediction decomposes future motion into a linear motion base, a self-motion bias, and a resonance bias \cite{wong_2025_resonance}, and produces predictions by superposing these components. Such a decomposition provides a meaningful structure for distinguishing individual and social factors in future motion.


However, existing methods mainly focus on how social context is encoded \cite{yuan_2021_agentformer}\cite{mohamed_2020_socialstgcnn}\cite{xu_2022_groupnet}, while paying less explicit attention to how the encoded social information is activated during future generation. Even with advanced encoders, the resulting social bias is typically injected via a fixed composition rule, inadvertently forcing it to exert a constant influence across all future steps.
We argue that dynamic encoding alone cannot guarantee dynamic activation. In reality, trajectory generation requires a continuous, context-aware trade-off between individual and social biases. Individual bias (e.g., physical inertia) dominates short-term predictions and sparse scenes, where over-activating social cues merely introduces noise. Conversely, dense scenes demand strong social activation to model complex interactions. Thus, fixed composition strategies inevitably under-activate essential social constraints in crowds, while over-activating noise in sparse or short-term scenarios.

To address this issue, we propose GEAR, a generation-aware bias activation model for human trajectory prediction. Instead of redesigning the social encoder, GEAR focuses on the final trajectory composition stage. It formulates future prediction as the composition of a linear motion base, a self-motion bias, and a social-resonance bias. Based on compact summaries of ego motion deviation and ego-neighbor resonance relations, GEAR generates step-wise activation gates for the self-motion and social-resonance bias branches. These gates dynamically modulate the two decoded bias terms before final trajectory composition, allowing the model to decide when and how strongly individual and social components should participate in generation. In this way, GEAR preserves the interpretable bias decomposition structure while introducing temporal flexibility into the generation process.

We evaluate GEAR on SDD \cite{andle_2023_stanford} and NBA \cite{wong_2025_resonance} benchmarks. Experimental results show that GEAR consistently improves the resonance-based baseline and achieves competitive state-of-the-art performance. Further analysis shows that the learned activation values vary across future prediction steps and scene densities, indicating that GEAR does not simply inject social information as a fixed condition. 

Our contributions are summarized as follows:

\begin{itemize}
    \item We identify the neglected gap between dynamic social encoding and dynamic social activation, and propose to explicitly control the activation strength of the individual and social components during trajectory generation.

    \item We propose GEAR, a generation-aware bias activation model. By employing a step-wise gating mechanism, GEAR dynamically modulates the self-motion and social-resonance biases at each future step, generating highly environment-adaptive trajectories.


    \item We evaluate GEAR on the ETH-UCY \cite{alahi_2016_social}, SDD \cite{andle_2023_stanford}, and NBA \cite{wong_2025_resonance} datasets, achieving state-of-the-art or competitive performance. Further analyses demonstrate its interpretable activation behaviors across various prediction horizons and scene densities.
\end{itemize}
\color{black}

\section{Related Work}

\subsection{Human Trajectory Prediction}

Human trajectory prediction aims to forecast future agent positions from observed motion histories. Early learning-based methods commonly employ recurrent neural networks to encode the target agent's historical trajectory and generate future positions autoregressively\cite{alahi_2016_social}\cite{salzmann_2020_trajectron}. Social-LSTM\cite{alahi_2016_social} introduces social pooling to aggregate hidden states of neighboring pedestrians, demonstrating the importance of interaction modeling in crowded scenes. Subsequent studies further improve trajectory prediction with generative modeling, endpoint- or waypoint-conditioned prediction, memory retrieval, and multimodal sampling\cite{gupta_2018_social}\cite{mangalam_2020_it}\cite{mangalam_2021_goals}\cite{xu_2022_remember}, thereby enhancing the diversity and accuracy of predicted trajectories. However, many of these methods still represent individual motion tendency and social influence within a coupled latent space.

Recent research has moved toward stronger spatio-temporal representation learning. Graph-based methods represent agents as nodes and interactions as edges, enabling structured message passing among neighboring agents\cite{mohamed_2020_socialstgcnn}\cite{xu_2022_groupnet}. Attention-based and Transformer-based models further allow each agent to selectively aggregate relevant temporal and social cues\cite{yuan_2021_agentformer}. Other methods exploit frequency-domain representations, diffusion models, or structured trajectory bases to characterize long-horizon uncertainty and multimodal futures\cite{xia_2025_another}\cite{liu_2024_uncertaintyaware}\cite{wong_2025_resonance}. Although these methods differ in architecture, most of them primarily focus on more effectively encoding individual motion and social context before generation. In contrast, this paper focuses on a complementary question: once social context or other motion factors have been encoded as components in trajectory generation, how should these components take effect during future trajectory composition?

\subsection{Social Interaction Modeling and Structured Trajectory Generation}

Social interaction modeling is a central problem in human trajectory prediction. Existing methods commonly capture the influence of neighboring agents through pooling, attention, graph message passing, scene-level interaction encoding, or learned social relation representations\cite{alahi_2016_social}\cite{mohamed_2020_socialstgcnn}\cite{yuan_2021_agentformer}\cite{wong_2024_socialcircle}. These mechanisms have become important designs in social trajectory forecasting and substantially improve the modeling of complex interactive scenes. However, stronger social context encoding does not necessarily mean that social information will be properly used during prediction. When interaction cues are weak or noisy, overusing social information may disturb the target agent's own motion continuity; when interaction constraints are strong, insufficient use of social information may lead to socially inconsistent forecasts.

Another line of work studies future trajectory prediction from the perspective of structured generation. These methods decompose future motion into several semantically different components, such as linear motion trends, goals or intentions, individual motion offsets, interaction responses, stochastic perturbations, or other residual terms\cite{mangalam_2020_it}\cite{mangalam_2021_goals}\cite{wong_2025_resonance}. Compared with compressing all motion variations into a single latent representation, structured trajectory generation provides a clearer explanation of motion formation and reduces the burden of learning multiple sources of variation simultaneously. Resonance-based trajectory prediction can be viewed as a representative example of this direction\cite{wong_2025_resonance}. It represents future trajectories as the composition of a base motion term, a self-motion bias term, and a social interaction correction term, providing a meaningful structure for distinguishing individual motion continuity from social interaction effects\cite{wong_2025_resonance}.

However, existing structured generation methods usually focus more on how to obtain these motion components, while paying less explicit attention to how these components should take effect at the final generation stage. Once different motion components are decoded, they are often directly combined into the final trajectory according to a fixed rule\cite{wong_2025_resonance}. Such static composition implicitly assumes that all types of motion components should participate in generation in the same manner across all future time steps, which is not always appropriate. Future trajectory generation often requires a dynamic trade-off between individual motion continuity and social interaction constraints according to prediction horizons and scene interaction intensity. Motivated by this issue, this paper studies dynamic component activation in structured trajectory generation.

\begin{figure*}[t]
    \centering
    \includegraphics[width=400pt]{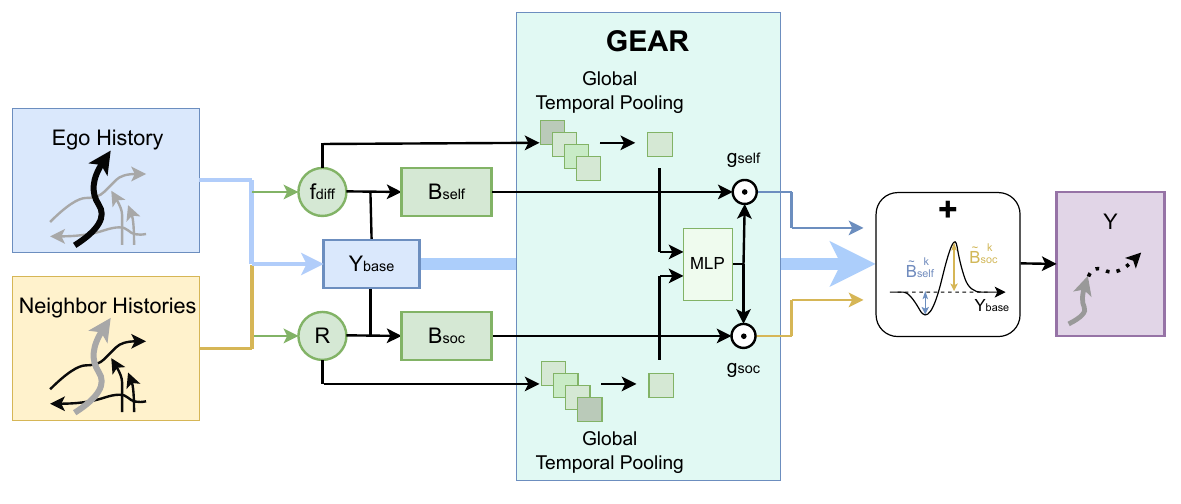}
    \caption{{The GEAR framework. A bias-decomposed trajectory generator equipped with a generation-aware activation mechanism for step-wise bias modulation.}}
    \label{fig:overview}
\end{figure*}

\subsection{Gating, Fusion, and Generation-Stage Modulation}

Gating and attention mechanisms have been widely used in trajectory prediction models to regulate interactions among different information sources. Existing methods commonly use such mechanisms to select relevant neighbors, aggregate social context, fuse heterogeneous features, control temporal state updates, or combine multimodal representations\cite{alahi_2016_social}\cite{yuan_2021_agentformer}\cite{wong_2024_socialcircle}. These designs mainly operate at the observation encoding stage or the intermediate feature fusion stage. Therefore, their core concern is usually what information should be extracted from historical trajectories and neighbor interactions, and how such information should be fused into the prediction representation.

Different from the above methods, this paper focuses on generation-stage modulation in structured trajectory generation. Even after a model has obtained several motion components with different semantic roles, it still needs to determine how these components should participate in final trajectory composition at each future time step. This problem differs from social information aggregation during encoding or intermediate feature fusion; instead, it is a dynamic calibration of the output composition process. Based on this perspective, GEAR performs step-wise activation on decoded bias components while preserving their semantic structure, thereby adaptively adjusting the contributions of different motion factors according to prediction horizons and scene interaction intensity.

\section{Methodology}

In this section, we present the proposed GEAR framework. The key idea of GEAR is to distinguish bias decoding from bias activation. Instead of directly adding decoded self-motion and social-resonance biases to the base trajectory with fixed strength, GEAR learns step-wise activation gates to determine how strongly each bias component should contribute to future generation.

\subsection{Problem Formulation}

Given the observed trajectory of a target agent $i$,
\begin{equation}
    \mathbf{X}_i = \{\mathbf{x}_i^1, \mathbf{x}_i^2, \ldots, \mathbf{x}_i^{T_{obs}}\},
\end{equation}
and the observed trajectories of its neighboring agents,
\begin{equation}
    \mathbf{X}_{-i} = \{\mathbf{X}_j \mid j \neq i\},
\end{equation}
the goal of trajectory prediction is to forecast the future trajectory of the target agent. Following the common multimodal trajectory prediction protocol, the model generates $K$ candidate future trajectories:
\begin{equation}
    \hat{\mathbf{Y}}_i^k =
    \{\hat{\mathbf{y}}_i^{k,1}, \hat{\mathbf{y}}_i^{k,2}, \ldots,
    \hat{\mathbf{y}}_i^{k,T_{pred}}\},
    \quad k=1,\ldots,K .
\end{equation}
During evaluation, we follow the best-of-$K$ protocol and report minADE$_K$/minFDE$_K$. In our experiments, we set $K=20$ to align with the evaluation protocol of the resonance-based backbone and prior trajectory prediction methods.

Most existing trajectory prediction models can be viewed as first encoding individual motion and social context from observations, and then decoding future positions from the learned representation. In contrast, our focus is not to redesign the social interaction encoder. Instead, we study how the decoded individual and social components should be activated during the final generation stage.

\subsection{Bias-Decomposed Generation and Its Limitation}

We build GEAR upon a bias-decomposed trajectory generation formulation. For the $k$-th sampled prediction of a target agent, the future trajectory is represented as the composition of three terms:
\begin{equation}
    \hat{\mathbf{Y}}_i^k(t)
    =
    \mathbf{Y}_{base,i}(t)
    +
    \mathbf{B}_{self,i}^k(t)
    +
    \mathbf{B}_{soc,i}^k(t),
    \label{eq:static_composition}
\end{equation}
where $\mathbf{Y}_{base,i}$ denotes the linear motion base, $\mathbf{B}_{self,i}^k$ denotes the sampled self-motion bias, and $\mathbf{B}_{soc,i}^k$ denotes the sampled social-resonance bias.

The linear motion base $\mathbf{Y}_{base,i}$ captures the coarse future motion trend by extrapolating the recent motion of the target agent. It serves as a reference trajectory around which nonlinear deviations are modeled. The self-motion bias $\mathbf{B}_{self,i}^k$ captures the target agent's individual nonlinear motion deviation from the linear base. It mainly reflects self-driven motion variations, such as intention changes or nonlinear continuation of the observed trajectory. The social-resonance bias $\mathbf{B}_{soc,i}^k$ captures interaction-induced corrections derived from ego-neighbor relations. It represents how surrounding agents affect the future motion of the target agent through social context.

This decomposition provides a natural structure for separating individual motion continuity and social interaction correction. However, the static composition in Eq.~\eqref{eq:static_composition} implicitly assumes that both bias terms are fully active at every future step:
\begin{equation}
    g_{self}(t)=1,\quad g_{soc}(t)=1,\quad
    \forall t \in \{1,\ldots,T_{pred}\}.
    \label{eq:static_gate}
\end{equation}
Such a fixed composition strategy may be suboptimal. In short-term prediction, the target agent's motion can be dominated by its own recent velocity and self-motion tendency, where overusing social correction may disturb local continuity. In dense scenes, social-resonance bias can provide useful constraints for modeling avoidance, following, and coordinated movement. In sparse scenes, however, the interaction evidence can be weak, and uniformly injecting social-resonance bias may introduce unnecessary corrections. Therefore, bias activation should be adaptive in both directions: strengthening social-resonance correction when interactions are strong, while suppressing excessive social influence when interaction cues are weak.

This motivates the central design of GEAR: decoded bias terms should not be treated as automatically active. Instead, their contributions should be dynamically modulated before final trajectory composition.

\subsection{Generation-Aware Bias Activation}

GEAR introduces a lightweight step-wise activation mechanism for bias composition. Instead of fusing features before decoding, GEAR first obtains the self-motion bias and social-resonance bias from the backbone, and then modulates the decoded bias terms at the final generation stage. Importantly, GEAR is orthogonal to the sampling process: it does not change how candidate self-motion and social-resonance biases are generated, but controls how the decoded biases are activated when composing each candidate trajectory.

Specifically, the activation module takes two compact context summaries as input. The first one summarizes the ego motion deviation from the linear base:
\begin{equation}
    \mathbf{c}_{self} = \mathrm{Pool}(\mathbf{f}_{diff}),
\end{equation}
where $\mathbf{f}_{diff}$ denotes the ego trajectory deviation feature relative to the linear motion base. The second one summarizes social resonance relations:
\begin{equation}
    \mathbf{c}_{soc} = \mathrm{Pool}(\mathbf{R}),
\end{equation}
where $\mathbf{R}$ denotes the resonance relation matrix encoded from ego-neighbor interactions. In our implementation, $\mathrm{Pool}(\cdot)$ is mean pooling over the corresponding temporal or relational dimension. The two summaries are concatenated to form the gate input:
\begin{equation}
    \mathbf{g}_{in} = [\mathbf{c}_{self}; \mathbf{c}_{soc}].
\end{equation}

The activation gates are generated by a two-layer MLP followed by a sigmoid function:
\begin{equation}
    [\mathbf{g}_{self}, \mathbf{g}_{soc}]
    =
    \sigma(\mathrm{MLP}(\mathbf{g}_{in})),
    \label{eq:gate_generation}
\end{equation}
where $\mathbf{g}_{self}, \mathbf{g}_{soc} \in \mathbb{R}^{T_{pred}}$. Each element $g_{self}(t)$ controls the activation strength of the self-motion bias at future step $t$, while $g_{soc}(t)$ controls the activation strength of the social-resonance bias at the same step. The sigmoid function constrains activation values into $[0,1]$, allowing GEAR to softly regulate the contribution of each bias branch.

The activation gates are shared across the $K$ sampled predictions, since they are generated from deterministic ego-motion and resonance summaries rather than sample-specific noise variables. For the $k$-th sampled prediction, the activated bias terms are computed as:
\begin{equation}
    \tilde{\mathbf{B}}_{self,i}^k(t)
    =
    g_{self}(t)\mathbf{B}_{self,i}^k(t),
    \label{eq:self_activation}
\end{equation}
\begin{equation}
    \tilde{\mathbf{B}}_{soc,i}^k(t)
    =
    g_{soc}(t)\mathbf{B}_{soc,i}^k(t).
    \label{eq:social_activation}
\end{equation}
The final prediction is then obtained by generation-aware bias composition:
\begin{equation}
    \hat{\mathbf{Y}}_i^k(t)
    =
    \mathbf{Y}_{base,i}(t)
    +
    \tilde{\mathbf{B}}_{self,i}^k(t)
    +
    \tilde{\mathbf{B}}_{soc,i}^k(t).
    \label{eq:gear_composition}
\end{equation}

Compared with static bias superposition, Eq.~\eqref{eq:gear_composition} explicitly models the activation process of decoded bias components. This design has two important properties. First, GEAR preserves the semantic decomposition of the trajectory generation process: the linear base, self-motion bias, and social-resonance bias remain separate components. Second, GEAR introduces temporal flexibility at the generation stage, enabling the model to adjust the relative contribution of self-motion and social interaction across future horizons. Different from feature-level fusion, GEAR does not mix the two bias branches before decoding, but modulates their step-wise activation before final composition.

The overall forward process of GEAR is summarized in
Algorithm~\ref{alg:gear_forward}. GEAR retains the original
bias decoding process and reuses the ego-motion deviation
feature and resonance relation representation to generate
step-wise activation gates. The gates are shared across the
$K$ sampled candidates and only modulate the decoded bias
terms before final trajectory composition.

\begin{algorithm}[t]
\caption{Forward Process of GEAR}
\label{alg:gear_forward}
\begin{algorithmic}[1]
\REQUIRE Ego history $\mathbf{X}_i$, neighbor histories
$\mathbf{X}_{-i}$, number of samples $K$
\ENSURE Candidate future trajectories
$\{\hat{\mathbf{Y}}_i^k\}_{k=1}^{K}$

\STATE $(\mathbf{f}_{diff}, \mathbf{X}_{linear},
\mathbf{Y}_{base}) \gets
\mathrm{LinearDiff}(\mathbf{X}_i)$

\STATE $\mathbf{R} \gets
\mathrm{Resonance}(\mathbf{X}_i,\mathbf{X}_{-i})$

\STATE $\mathbf{c}_{self} \gets
\mathrm{MeanPool}(\mathbf{f}_{diff})$

\STATE $\mathbf{c}_{soc} \gets
\mathrm{MeanPool}(\mathbf{R})$

\STATE $[\mathbf{g}_{self},\mathbf{g}_{soc}]
\gets
\sigma\!\left(
\mathrm{MLP}([\mathbf{c}_{self};\mathbf{c}_{soc}])
\right)$

\FOR{$k=1$ to $K$}
    \STATE $\mathbf{B}_{self}^{k}
    \gets
    \mathrm{SelfBias}(\mathbf{X}_{linear},
    \mathbf{f}_{diff};\mathbf{z}_{self}^{k})$

    \STATE $\mathbf{B}_{soc}^{k}
    \gets
    \mathrm{SocialBias}(
    \mathbf{X}_i-\mathbf{X}_{linear},
    \mathbf{f}_{diff},\mathbf{R};
    \mathbf{z}_{soc}^{k})$

    \STATE $\widetilde{\mathbf{B}}_{self}^{k}
    \gets
    \mathbf{g}_{self}\odot\mathbf{B}_{self}^{k}$

    \STATE $\widetilde{\mathbf{B}}_{soc}^{k}
    \gets
    \mathbf{g}_{soc}\odot\mathbf{B}_{soc}^{k}$

    \STATE $\hat{\mathbf{Y}}_i^{k}
    \gets
    \mathbf{Y}_{base}
    +
    \widetilde{\mathbf{B}}_{self}^{k}
    +
    \widetilde{\mathbf{B}}_{soc}^{k}$
\ENDFOR

\RETURN $\{\hat{\mathbf{Y}}_i^k\}_{k=1}^{K}$
\end{algorithmic}
\end{algorithm}

\subsection{Training Objective}

Following the backbone training protocol, GEAR is trained end-to-end with a best-of-$K$ trajectory regression objective. Given the ground-truth future trajectory $\mathbf{Y}_i$ and $K$ predicted trajectories $\{\hat{\mathbf{Y}}_i^k\}_{k=1}^{K}$, the loss is defined as
\begin{equation}
    \mathcal{L}_{traj}
    =
    \min_{k \in \{1,\ldots,K\}}
    \frac{1}{T_{pred}}
    \sum_{t=1}^{T_{pred}}
    \left\|
    \hat{\mathbf{Y}}_i^k(t)-\mathbf{Y}_i(t)
    \right\|_2.
    \label{eq:loss}
\end{equation}
The model is trained under the same sampling and optimization protocol as the resonance-based backbone. The final output of the model is a set of candidate future trajectory coordinates, while internally each trajectory is represented as the composition of a linear base and activated residual bias terms.

\section{Experiments}

In this section, we evaluate the proposed GEAR on standard trajectory prediction benchmarks. We first describe the experimental settings, then compare GEAR with state-of-the-art methods, followed by ablation studies and activation analyzes.

\subsection{Experimental Setup}

\textbf{Datasets.}
We conduct experiments on three widely used trajectory prediction benchmarks: ETH-UCY \cite{alahi_2016_social}, Stanford Drone Dataset (SDD) \cite{andle_2023_stanford}, and NBA \cite{wong_2025_resonance}. ETH-UCY contains pedestrian trajectories collected from five real-world scenes, including ETH, HOTEL, UNIV, ZARA1, and ZARA2. Following the standard leave-one-out protocol, models are trained on four scenes and tested on the remaining one. SDD contains large-scale trajectories captured from a university campus, covering diverse agent categories and interaction patterns. NBA contains player trajectories from basketball games, where motion prediction is strongly affected by team-level coordination and opponent interactions. Following prior works, we report the results on NBA under two prediction horizons, $t_f=5$ and $t_f=10$.

\textbf{Metrics.}
We use Average Displacement Error (ADE) \cite{chen_2021_human} and Final Displacement Error (FDE) \cite{chen_2021_human} as evaluation metrics. Following the evaluation protocol of Resonance and previous multimodal trajectory prediction methods, the model generates $K=20$ candidate trajectories, and the best prediction is selected according to the displacement error. For brevity, we denote minADE$_{20}$/minFDE$_{20}$ as ADE/FDE in all tables. Lower values indicate better performance.

\textbf{Implementation Details.}
GEAR is implemented upon the Resonance trajectory generation
framework while retaining its original linear-base, self-bias,
and social-resonance bias branches. Rather than modifying the
original bias encoding and decoding process, GEAR reuses the
ego-motion deviation feature $\mathbf{f}_{diff}$ and the
social-resonance relation representation $\mathbf{R}$ produced
by the backbone to generate step-wise activation gates for final
trajectory composition.

Specifically, mean pooling is applied to $\mathbf{f}_{diff}$
and $\mathbf{R}$ along their corresponding temporal or
relational dimensions. The resulting summaries are concatenated
and passed through a two-layer MLP, which outputs a vector of
length $2T_{pred}$. The output is divided into the self-bias
gate $\mathbf{g}_{self}$ and the social-bias gate
$\mathbf{g}_{soc}$, both of which are constrained to $[0,1]$
using a sigmoid function. At each future step, the corresponding
gate value is broadcast over the trajectory coordinate
dimensions and used to scale the decoded self-motion or
social-resonance bias. Since the gates are generated from
deterministic motion and interaction summaries, all $K$
candidate trajectories of the same input share the same gate
sequences. Therefore, GEAR does not alter the multimodal
sampling process or the number of trajectory candidates.

For all datasets, we follow the data splits, coordinate
preprocessing, and evaluation protocols of Resonance to ensure
a fair comparison. During evaluation, the model generates
$K=20$ candidate trajectories and reports best-of-20 ADE/FDE.
All experiments are conducted on an NVIDIA RTX 4090 GPU.

\begin{table*}[t]
  \caption{Comparison with state-of-the-art methods on ETH-UCY~\cite{alahi_2016_social} and SDD~\cite{andle_2023_stanford}. Metrics are ADE/FDE under best-of-20 evaluation. Results are reported in meters on ETH-UCY and in pixels on SDD. Lower is better. The top2 results are highlighted in bold and the best results are bold and underlined.}
  \label{tab:crowd_benchmarks}
  \centering
  \footnotesize
  \setlength{\tabcolsep}{4pt}
  \renewcommand{\arraystretch}{1.08}
  \begin{minipage}[t]{0.72\textwidth}
    \begin{tabular*}{\linewidth}{@{\extracolsep{\fill}}lcccccc@{}}
      \toprule
      Method (ETH-UCY) & eth$\downarrow$ & hotel$\downarrow$ & univ$\downarrow$ & zara1$\downarrow$ & zara2$\downarrow$ & Average$\downarrow$ \\
      \midrule
      MS-TIP~\cite{chib_2024_mstip} (2024) & 0.39 / 0.57 & 0.13 / 0.22 & 0.24 / 0.40 & 0.20 / 0.34 & 0.17 / 0.29 & 0.22 / 0.36 \\
      SMEMO~\cite{marchetti_2024_smemo} (2024) & 0.39 / 0.59 & 0.14 / 0.20 & 0.23 / 0.41 & 0.19 / 0.32 & 0.15 / 0.26 & 0.22 / 0.35 \\
      Trajectron++~\cite{salzmann_2020_trajectron} (2020) & 0.43 / 0.86 & 0.12 / 0.19 & 0.22 / 0.43 & 0.17 / 0.32 & \textbf{0.12} / 0.25 & 0.20 / 0.39 \\
      LG-Traj~\cite{chib_2025_lgtraj} (2024) & 0.38 / 0.56 & \textbf{0.11} / 0.17 & 0.23 / 0.42 & 0.18 / 0.33 & 0.14 / 0.25 & 0.20 / 0.34 \\
      PPT~\cite{lin_2024_progressive} (2024) & 0.36 / 0.51 & \textbf{0.11} / \underline{\textbf{0.14}} & 0.22 / 0.40 & 0.17 / 0.30 & \textbf{0.12} / \textbf{0.21} & 0.20 / 0.31 \\
      E-V$^2$-Net~\cite{xia_2025_another} (2025) & 0.25 / 0.38 & \textbf{0.11} / 0.16 & 0.23 / 0.42 & 0.19 / 0.30 & 0.13 / 0.24 & 0.18 / 0.30 \\
      AgentFormer~\cite{yuan_2021_agentformer} (2021) & 0.26 / 0.39 & \textbf{0.11} / \underline{\textbf{0.14}} & 0.26 / 0.46 & \underline{\textbf{0.15}} / \underline{\textbf{0.23}} & 0.14 / 0.23 & 0.18 / 0.29 \\
      SocialCircle~\cite{wong_2024_socialcircle} (2024) & 0.25 / 0.38 & 0.12 / \underline{\textbf{0.14}} & 0.23 / 0.42 & 0.18 / 0.29 & 0.13 / 0.22 & 0.18 / 0.29 \\
      Y-net~\cite{mangalam_2021_goals} (2021) & 0.28 / \underline{\textbf{0.33}} & \underline{\textbf{0.10}} / \underline{\textbf{0.14}} & 0.24 / 0.41 & 0.17 / 0.27 & 0.13 / 0.22 & 0.18 / \underline{\textbf{0.27}} \\
      UPDD~\cite{liu_2024_uncertaintyaware} (2024) & \underline{\textbf{0.22}} / 0.42 & 0.17 / 0.30 & \underline{\textbf{0.14}} / \underline{\textbf{0.28}} & 0.16 / 0.30 & 0.14 / 0.31 & 0.17 / 0.32 \\
      TAMLD~\cite{ren_2025_totp} (2025) & 0.39 / 0.58 & 0.13 / 0.18 & 0.22 / 0.37 & 0.18 / 0.28 & 0.13 / \textbf{0.21} & 0.21 / 0.32 \\
      Resonance~\cite{wong_2025_resonance} (2025) & \textbf{0.23} / \textbf{0.35} & \underline{\textbf{0.10}} / \textbf{0.15} & 0.24 / 0.41 & 0.17 / 0.29 & 0.13 / 0.22 & 0.17 / \textbf{0.28} \\
      IAD~\cite{liu_2026_intentionaware} (2026) & 0.34 / 0.52 & 0.15 / 0.24 & \textbf{0.20} / \textbf{0.36} & \textbf{0.15} / \textbf{0.24} & \underline{\textbf{0.11}} / \underline{\textbf{0.20}} & 0.19 / 0.31 \\
      \midrule
      GEAR (Ours) & \underline{\textbf{0.22}} / \textbf{0.35} & \underline{\textbf{0.10}} / 0.16 & 0.23 / 0.41 & \textbf{0.16} / 0.28 & \textbf{0.12} / 0.22 & \underline{\textbf{0.16}} / \textbf{0.28} \\
      \bottomrule
    \end{tabular*}
  \end{minipage}%
  \hspace{0.008\textwidth}%
  \begin{minipage}[t]{0.24\textwidth}
    \begin{tabular*}{\linewidth}{@{\extracolsep{\fill}}lc@{}}
      \toprule
      Method (SDD) & ADE/FDE$\downarrow$ \\
      \midrule
      FlowChain~\cite{maeda_2023_fast} (2023) & 9.93 / 17.17 \\
      IMP~\cite{shi_2023_representing} (2023) & 8.98 / 15.54 \\
      SMEMO~\cite{marchetti_2024_smemo} (2024) & 8.11 / 13.06 \\
      LG-Traj~\cite{chib_2025_lgtraj} (2024) & 7.80 / 12.79 \\
      Y-net~\cite{mangalam_2021_goals} (2021) & 7.85 / 11.85 \\
      UEN~\cite{su_2024_unified} (2024) & 7.30 / 10.40 \\
      PPT~\cite{lin_2024_progressive} (2024) & 7.03 / 10.65 \\
      UPDD~\cite{liu_2024_uncertaintyaware} (2024) & 6.59 / 13.90 \\
      E-V$^2$-Net~\cite{xia_2025_another} (2025) & 6.57 / 10.49 \\
      SocialCircle~\cite{wong_2024_socialcircle} (2024) & 6.54 / 10.36 \\
      MUSE-VAE~\cite{lee_2022_musevae} (2022) & 6.36 / 11.10 \\
      IAD~\cite{liu_2026_intentionaware} (2026) & 6.85 / 11.22 \\
      Resonance~\cite{wong_2025_resonance} (2025) & \underline{\textbf{6.27}} / \underline{\textbf{10.02}} \\
      \midrule
      GEAR (Ours) & \underline{\textbf{6.18}} / \underline{\textbf{9.89}} \\
      \bottomrule
    \end{tabular*}
  \end{minipage}
\end{table*}




\subsection{Comparison with State-of-the-Art Methods}

We first compare GEAR with existing state-of-the-art methods
on the ETH-UCY and SDD benchmarks. As shown in
Table~\ref{tab:crowd_benchmarks}, GEAR achieves strong
performance on both datasets. On ETH-UCY, GEAR improves the
average ADE of the Resonance baseline from $0.17$ to $0.16$
while maintaining the same average FDE of $0.28$. It also
achieves competitive results across the five leave-one-out
scenes, indicating that the improvement is not driven by a
single test split. These results suggest that step-wise bias
activation can refine trajectory composition without disrupting
the strong generalization ability of the original
bias-decomposed predictor.

The advantage is more evident on SDD, which contains diverse
agent categories, large-scale scenes, and complex interaction
patterns. GEAR improves the Resonance baseline from
$6.27/10.02$ to $6.18/9.89$ in ADE/FDE and achieves the best
performance among the compared methods. Since the original
backbone already models ego--neighbor resonance relations,
this additional gain indicates that improving social
representation alone is not sufficient. Explicitly calibrating
how the decoded self-motion and social-resonance components
participate in final trajectory generation provides an
additional and complementary benefit.

We further evaluate GEAR on the NBA dataset to examine
whether the proposed activation mechanism remains effective
beyond pedestrian and campus scenarios. As shown in
Table~\ref{tab:nba_results}, GEAR improves the already strong
Resonance baseline under both prediction horizons. For
$t_f=5$, GEAR reduces FDE from $0.78$ to $0.77$ while
maintaining the best ADE of $0.60$. For $t_f=10$, it improves
ADE/FDE from $1.12/1.38$ to $1.11/1.37$. Although the
numerical margins are modest, the improvements are obtained
consistently over a strong baseline under both short- and
longer-horizon settings.

\begin{table}[t]
\centering
\caption{Comparison on NBA. Metrics are ADE/FDE under best-of-20 evaluation. Lower is better.}
\label{tab:nba_results}

\setlength{\tabcolsep}{9pt}
\renewcommand{\arraystretch}{1.10}
\resizebox{\columnwidth}{!}{
\begin{tabular}{lcc}
\toprule
\textbf{Method} & $t_f = 5 \downarrow$ & $t_f = 10 \downarrow$ \\
\midrule
PECNet~\cite{mangalam_2020_it}
& 0.96 / 1.69 & 1.83 / 3.41 \\

MemoNet~\cite{xu_2022_remember}
& 0.71 / 1.14 & 1.25 / 1.47 \\

GroupNet+NMMP~\cite{xu_2022_groupnet}
& 0.69 / 1.08 & 1.25 / 1.80 \\

GroupNet+CVAE~\cite{xu_2022_groupnet}
& 0.62 / 0.95 & 1.13 / 1.69 \\

V$^2$-Net~\cite{wong_2022_view}
& 0.69 / 0.96 & 1.28 / 1.68 \\

E-V$^2$-Net~\cite{xia_2025_another}
& 0.68 / 0.93 & 1.26 / 1.64 \\

SocialCircle~\cite{wong_2024_socialcircle}
& 0.67 / 0.90 & 1.18 / 1.46 \\

SocialCircle+~\cite{wong_2024_socialcirclea}
& 0.65 / 0.86 & 1.14 / \textbf{1.37} \\

Resonance~\cite{wong_2025_resonance}
& \textbf{0.60} / 0.78 & 1.12 / 1.38 \\
\midrule
\textbf{GEAR (Ours)}
& \textbf{0.60} / \textbf{0.77}
& \textbf{1.11} / \textbf{1.37} \\
\bottomrule
\end{tabular}
}
\end{table}

We further evaluate GEAR on another trajectory prediction framework to examine its applicability beyond the original backbone. For architectures that can disentangle personal and social information into explicit and semantically stable representations before trajectory decoding, GEAR more readily forms a natural match: such a decomposition provides a direct and clear interface for generation-stage activation. We apply GEAR to NSP \cite{NSP}(An alternative structured trajectory prediction framework). After incorporating GEAR, ADE remains unchanged at 6.52, while FDE decreases slightly from 10.52 to 10.51, indicating that its gains are relatively limited. We attribute this result to the structural difference between NSP and the original framework: although NSP also contains an independent component decomposition, its components do not correspond as explicitly to personal motion factors and social motion factors as those in the original framework. As a result, the generation-stage activation mechanism is less aligned with these semantics. More broadly, many trajectory prediction methods do not explicitly represent such decomposable motion components before decoding, which also makes it difficult for the advantages of GEAR to be fully realized. Nevertheless, for structured generators that explicitly represent motion components during trajectory generation, GEAR can be understood as a general generation-stage activation principle, rather than merely a mechanism tailored to a specific backbone.

Overall, the results across ETH-UCY, SDD, and NBA show that
GEAR is not tied to a specific scene type or motion domain.
The proposed activation mechanism improves trajectory
generation in pedestrian crowds, heterogeneous campus scenes,
and highly interactive sports scenarios. This cross-dataset
consistency supports our central claim that the contribution of
decoded motion components should be dynamically calibrated
during generation rather than combined through a fixed
composition rule.


\begin{figure*}[t]
    \centering
    \begin{subfigure}[t]{0.48\textwidth}
        \centering
        \includegraphics[width=\linewidth]{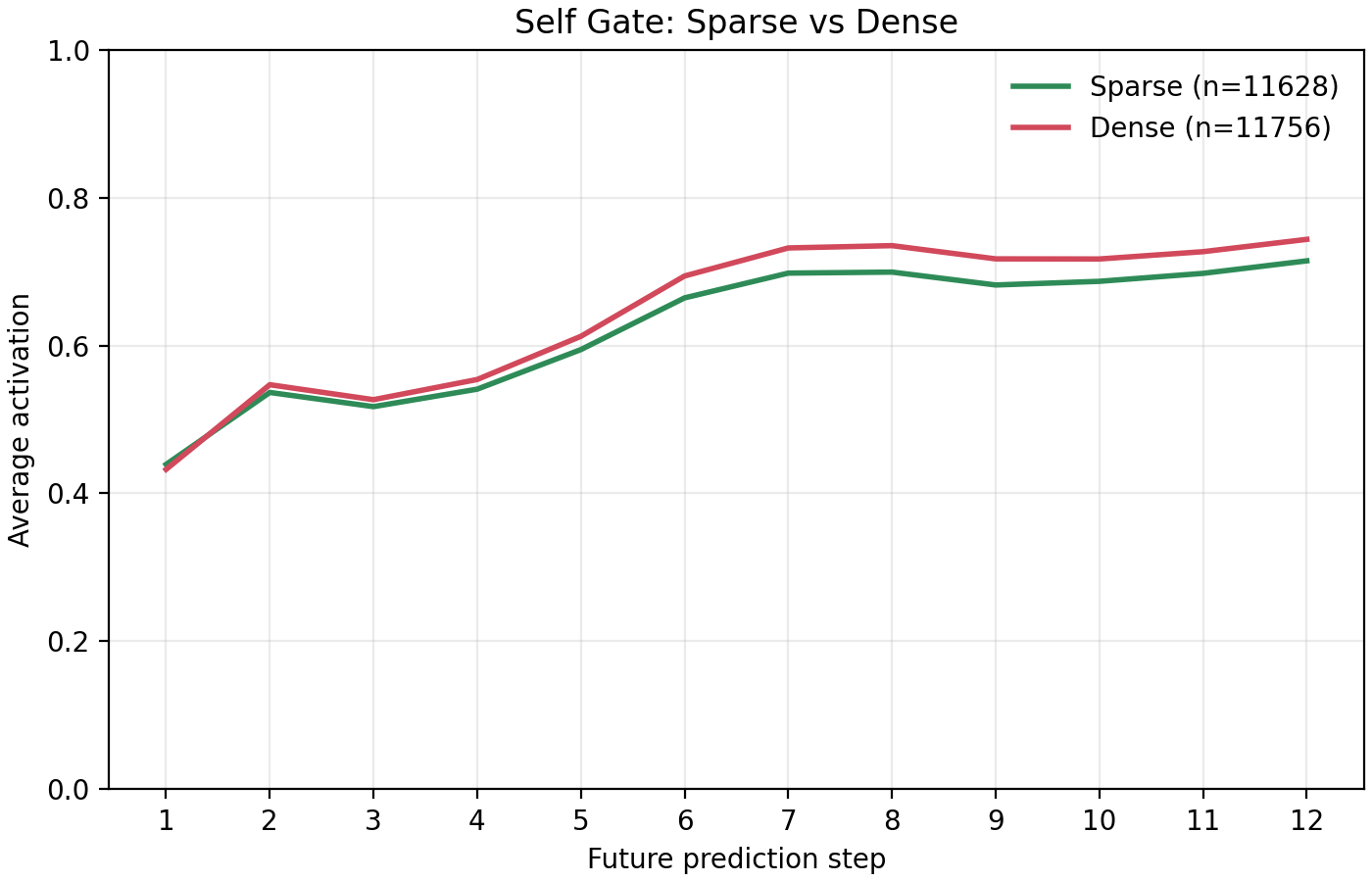}
        \caption{Self-bias activation under sparse and dense scenes.}
        \label{fig:self_bias_activation}
    \end{subfigure}
    \hfill
    \begin{subfigure}[t]{0.48\textwidth}
        \centering
        \includegraphics[width=\linewidth]{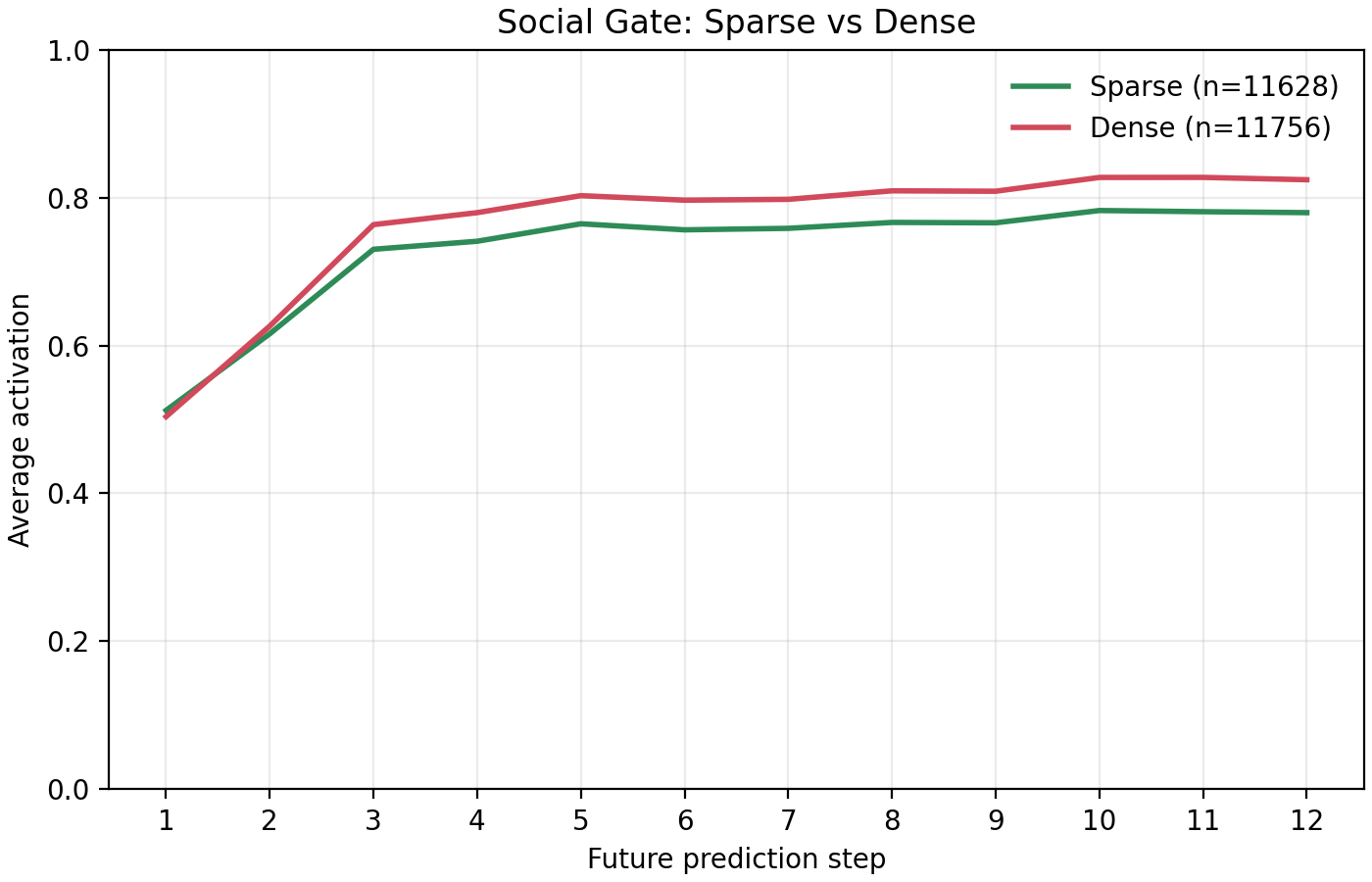}
        \caption{Social-bias activation under sparse and dense scenes.}
        \label{fig:social_bias_activation}
    \end{subfigure}
    \caption{Generation-aware activation patterns learned by GEAR. The model exhibits time-varying activation for self-motion and social-resonance biases, while dense scenes induce stronger social activation.}
    \label{fig:activation_patterns}
\end{figure*}

\begin{figure}[t]
    \centering
    \includegraphics[width=\linewidth]{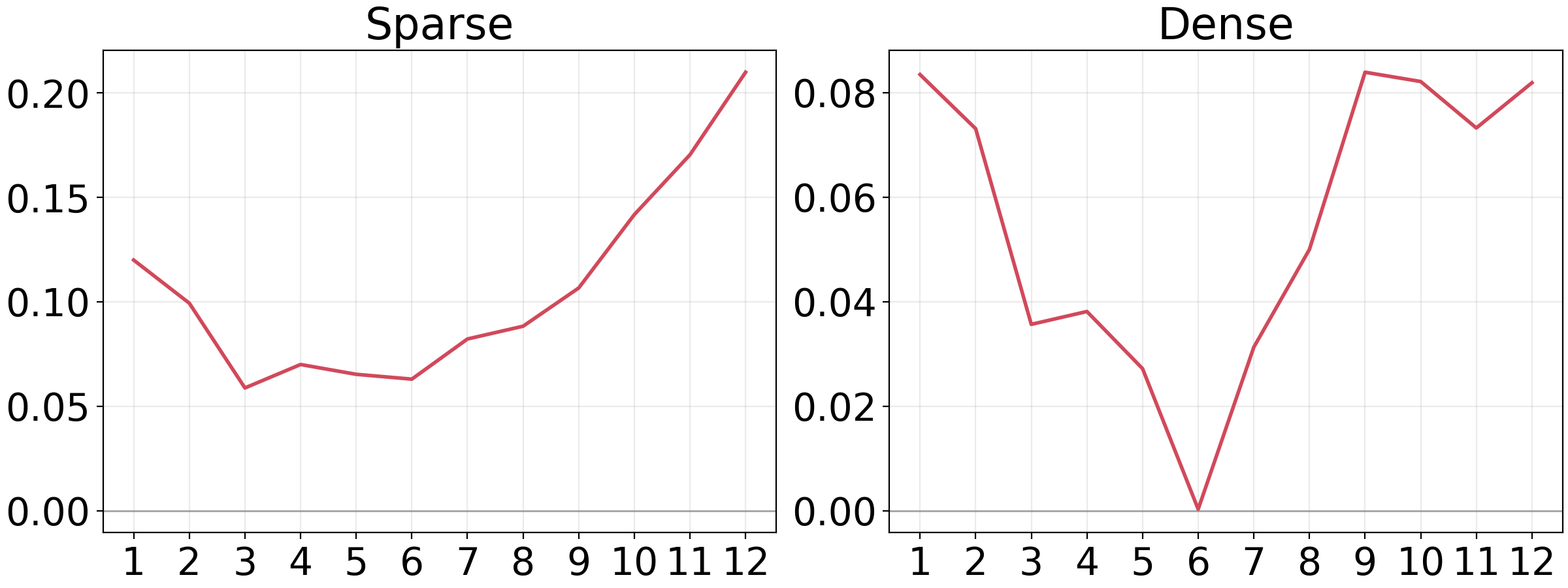}
    \caption{Step-wise error reduction of GEAR over Resonance in sparse and dense scenes. Positive values indicate lower errors achieved by GEAR. The gains are larger in sparse scenes and increase toward longer prediction horizons.}
    \label{fig:stepwise_error}
\end{figure}

\subsection{Ablation Study}

We conduct ablation studies to validate the design of generation-aware bias activation. The results are shown in Table~\ref{tab:ablation}. Starting from the Resonance baseline, adding only the social-bias activation gate already improves performance, indicating that the contribution of social-resonance bias should not be uniformly injected across future steps. The full GEAR model further introduces the self-bias activation gate and achieves the best performance, suggesting that future trajectory generation requires coordinated activation of both individual-motion and social-resonance biases.


\begin{table}[t]
\centering
\caption{Ablation study on bias activation design. Metrics are ADE/FDE under best-of-20 evaluation. Lower is better.}
\label{tab:ablation}

\setlength{\tabcolsep}{9pt}
\renewcommand{\arraystretch}{1.15}
\resizebox{0.88\columnwidth}{!}{
\begin{tabular}{lcc}
\toprule
Variant & ADE $\downarrow$ & FDE $\downarrow$ \\
\midrule
Baseline             & 6.27 & 10.01 \\
+ Social-bias Gate   & 6.20 & 9.92  \\
Attention Fusion     & 6.18 & 9.92  \\
Mean+Max Pooling     & 6.19 & 9.91  \\
\textbf{GEAR Full}   & \textbf{6.18} & \textbf{9.89} \\
\bottomrule
\end{tabular}
}
\end{table}

We also compare GEAR with two stronger but less structured variants. The attention-based fusion variant replaces the proposed step-wise bias activation with a heavier attention-based composition module. The mean+max pooling variant uses both mean and max pooling to summarize the context for gate generation. Both variants perform worse than the proposed design. This suggests that the improvement of GEAR does not come from simply increasing the fusion complexity. Instead, preserving the decomposed roles of self-motion and social-resonance biases and learning lightweight step-wise activation is more effective for trajectory generation.

\paragraph{Computational complexity.}
We further compare the model complexity of GEAR with the Resonance baseline on SDD. As shown in Table~\ref{tab:complexity}, GEAR introduces only a negligible computational overhead. The number of parameters increases from 3.15M to 3.18M, while the FLOPs increase from 31.44G to 31.48G. This confirms that the performance improvement mainly comes from generation-aware bias activation rather than substantially increased model capacity.

\begin{table}[t]
\centering
\caption{Model complexity on SDD. Parameters are reported in millions (M), and FLOPs are reported in billions (G).}
\label{tab:complexity}
\setlength{\tabcolsep}{10pt}
\renewcommand{\arraystretch}{1.05}
\begin{tabular}{lcc}
\toprule
Method & Params (M) & FLOPs (G) \\
\midrule
Resonance(Baseline) & 3.15 & 31.44 \\
GEAR & 3.18 & 31.48 \\
\bottomrule
\end{tabular}
\end{table}

\begin{table}[t]
\centering
\caption{Performance under different scene densities. Metrics are step-wise ADE/FDE averaged over samples in each group. Gain denotes the improvement over Resonance.}
\label{tab:density_gain}

\renewcommand{\arraystretch}{1.12}
\resizebox{0.98\columnwidth}{!}{
\begin{tabular}{lccc}
\toprule
Density & Baseline & GEAR & Gain \\
\midrule
Sparse
& 7.052 / 13.595
& 6.930 / 13.408
& 1.8\% / 1.4\% \\

Medium
& 5.198 / 10.000
& 5.146 / 9.931
& 1.0\% / 0.7\% \\

Dense
& 6.889 / 12.913
& 6.811 / 12.819
& 1.1\% / 0.7\% \\
\bottomrule
\end{tabular}
}
\end{table}

\begin{figure*}[t]
    \centering
    \includegraphics[width=\textwidth]{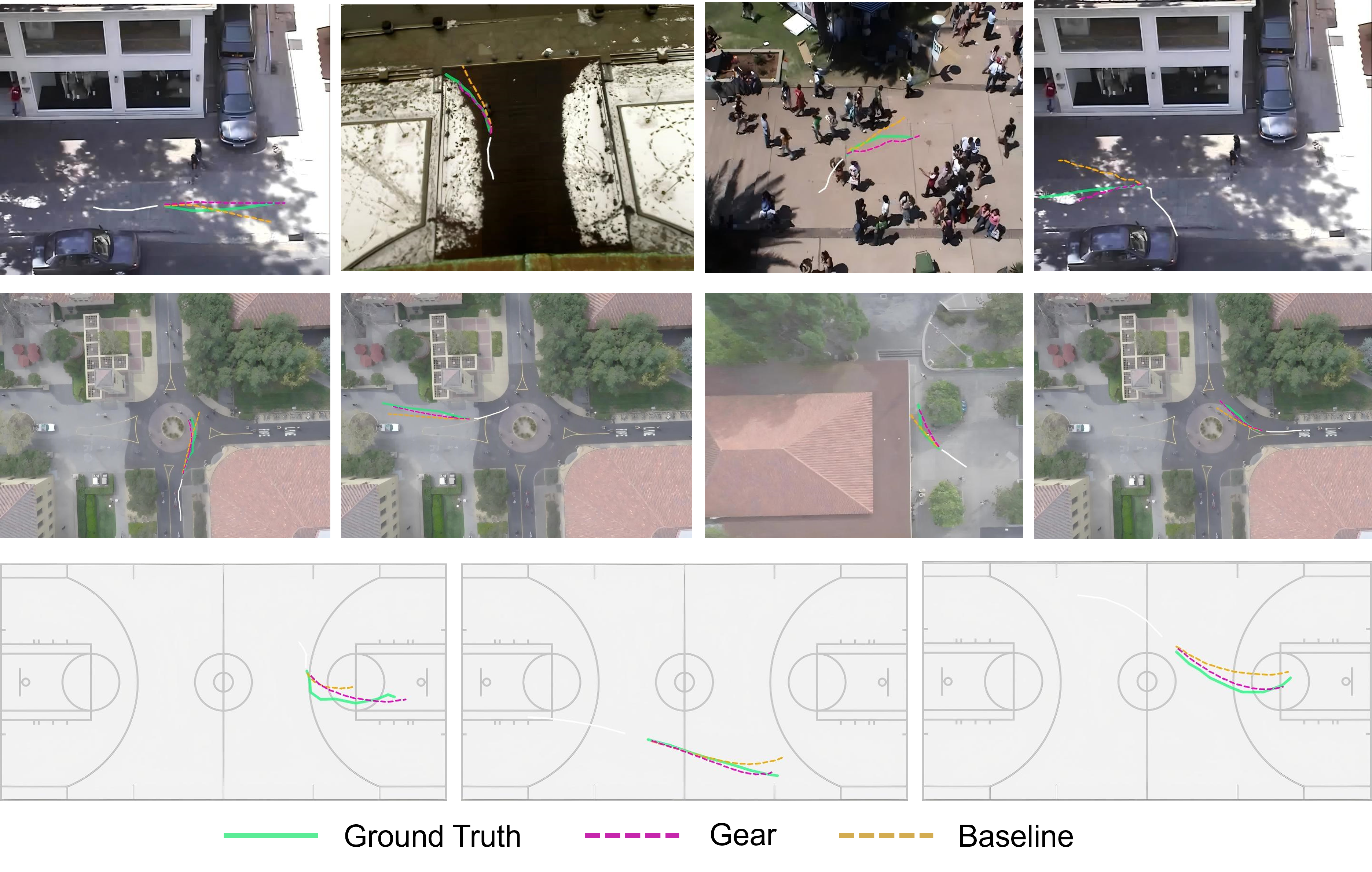}
    \caption{
    Qualitative comparison between Resonance and GEAR on ETH-UCY, SDD, and NBA.
    Each case shows the observed trajectory, ground-truth future trajectory, Resonance prediction, and GEAR prediction.
    Resonance uses static bias composition, whereas GEAR dynamically activates the decoded bias components during trajectory generation.
    GEAR produces predictions that more closely follow the ground truth across both crowd and sports scenarios.
    }
    \label{fig:qualitative_results}
\end{figure*}


\subsection{Analysis of Generation-Aware Activation}

To understand how GEAR calibrates decoded bias components, we analyze its learned activation values and prediction errors under different scene densities. We divide test samples into sparse, medium and dense groups according to the number of neighboring agents, using a quantile-based split to avoid severe sample imbalance. Figure~\ref{fig:activation_patterns} visualizes the activation curves of the sparse and dense groups, while Table~\ref{tab:density_gain} and Figure~\ref{fig:stepwise_error} report density-grouped and step-wise prediction improvements, respectively.


As shown in Figure~\ref{fig:activation_patterns}, both self-bias and social-bias activations vary substantially across future steps, confirming decoded components are not injected with fixed strengths. The activation values are relatively low at the beginning of prediction and increase as the prediction horizon extends. This pattern is consistent with the role of the linear motion base: near-future motion is largely determined by recent velocity and is therefore well approximated by $Y_{\mathrm{base}}$, while longer-horizon prediction increasingly requires nonlinear self-motion and interaction-aware corrections.

The benefit is particularly evident in sparse scenes. As reported in Table~\ref{tab:density_gain}, GEAR achieves the largest relative improvement in this group, and Figure~\ref{fig:stepwise_error} shows that the gain becomes increasingly pronounced with the following prediction steps. When interaction evidence is weak, the static composition of Resonance still injects the decoded social-resonance bias with full strength at every future step. Such premature or unnecessary corrections can interfere with the motion trend already captured by $Y_{\mathrm{base}}$ and may accumulate into larger long-horizon deviations. GEAR instead keeps the social-bias activation lower in sparse scenes, allowing early predictions to rely more on the linear base and introducing nonlinear corrections progressively when they become necessary.

Dense scenes exhibit consistently higher self-bias and social-bias activations, indicating that GEAR preserves stronger residual corrections when interaction constraints are more explicit. Notably, the self-bias branch also becomes more active, suggesting that socially complex motion is modeled by coordinating the two components. The positive step-wise gains in dense scenes further show that this calibration remains beneficial under strong interactions. Overall, GEAR plays a dual role: it suppresses unnecessary bias injection when social evidence is weak, while retaining stronger self-motion and social-resonance corrections when interaction constraints are substantial.


\begin{table}[t]
\centering
\caption{Prediction performance across different horizons on SDD.
Metrics are ADE/FDE, and Gain denotes the relative improvement
of GEAR over the baseline. Lower is better.}
\label{tab:horizon_results}

\setlength{\tabcolsep}{9pt}
\renewcommand{\arraystretch}{1.15}
\resizebox{0.98\columnwidth}{!}{
\begin{tabular}{lccc}
\toprule
Horizon & Baseline & GEAR & Gain (\%) \\
\midrule
Short  & 2.39 / 3.24  & \textbf{2.27 / 3.19} & 5.02 / 1.48 \\
Middle & 4.16 / 7.11  & \textbf{4.09 / 7.11} & 1.88 / 0.08 \\
Full   & 6.27 / 10.01 & \textbf{6.18 / 9.89} & 1.44 / 1.20 \\
\bottomrule
\end{tabular}
}
\end{table}

We further evaluate GEAR under short-, middle-, and full-horizon
settings, as reported in Table~\ref{tab:horizon_results}. GEAR
consistently improves the baseline across different prediction
horizons, demonstrating that generation-aware activation is
effective throughout the trajectory generation process. The
pronounced improvement in short-horizon ADE indicates that GEAR
can better calibrate decoded bias components at the early generation
stage, avoiding premature deviations from the motion trend captured
by the linear base. The gains remain evident over the full prediction
horizon, showing that step-wise activation also helps reduce
accumulated trajectory errors. Overall, these results confirm that
GEAR provides complementary benefits at different prediction
stages rather than being limited to a particular horizon.

\subsection{Qualitative Results}

We further visualize representative prediction cases from ETH-UCY, SDD, and NBA to qualitatively compare GEAR with the resonance-based baseline. As shown in Fig.~\ref{fig:qualitative_results}, each case includes the observed history, the future trajectory of ground-truth, the prediction of Resonance with static bias composition, and the prediction of GEAR with generation-aware bias activation.

Across all three datasets, GEAR produces trajectories that better match the ground truth while preserving plausible motion continuity. In crowd scenarios from ETH-UCY and SDD, the resonance-based baseline may either deviate too early or fail to introduce sufficient interaction-aware correction when local social constraints become important. In contrast, GEAR yields smoother and more adaptive predictions by dynamically modulating the contributions of self-motion and social-resonance biases across future steps. Similar behavior is observed on NBA, where GEAR better captures long-horizon motion changes under strong player interactions. These qualitative results provide intuitive evidence that generation-aware bias activation improves final trajectory composition beyond static bias superposition.

\section{Conclusion}

In this paper, we presented GEAR, a generation-aware bias activation model for social trajectory prediction. We revisited the bias-decomposed trajectory generation process and identified a missing activation problem: dynamically encoding social context does not necessarily ensure that the encoded social information is properly activated during future generation. To address this issue, GEAR preserves the decomposed structure of a linear motion base, a self-motion bias, and a social-resonance bias, while introducing step-wise activation gates to modulate the contribution of self-motion and social-resonance components before final trajectory composition. Experiments on ETH-UCY, SDD, and NBA demonstrate that GEAR consistently improves the resonance-based baseline and achieves competitive state-of-the-art performance. Further ablation studies, density-grouped analyses, activation visualizations, and qualitative results validate that generation-aware activation provides a more flexible and interpretable way to balance individual motion continuity and social interaction constraints. These results suggest that future trajectory prediction models should consider not only how social context is encoded, but also when it should be strengthened and when it should be suppressed during trajectory generation.

\section*{Acknowledgment}

This work was supported by the National Key Research and Development Program of China (2024YFF0617702), the National Natural Science Foundation of China (62502082, U22A2025, 62232007, U23A20309), the Fundamental Research Funds for the Central Universities (N25XQD014), and the 111 Project (B16009).

\bibliographystyle{acm}
\bibliography{ref}





\end{document}